\documentclass[a4paper]{llncs}
\usepackage[T1]{fontenc}
\usepackage[utf8]{inputenc}
\usepackage{floatrow}
\newfloatcommand{capbtabbox}{table}[][\FBwidth]
\usepackage[marginal]{footmisc}

\usepackage{authblk}
\usepackage{graphicx}
\usepackage{amsmath}
\usepackage{bm}
\usepackage{float}
\usepackage{makeidx}  
\usepackage{amssymb}
\usepackage{verbatim}
\usepackage{algorithm}
\usepackage{algorithmic}
\usepackage{booktabs}
\usepackage{multirow}
\usepackage{pbox} 
\usepackage{cite}
\usepackage{threeparttable}
\usepackage{caption}
\usepackage{subcaption}
\usepackage{hyperref}
\usepackage{makecell}
\usepackage{booktabs}
\usepackage{multirow}
\usepackage{caption}
\usepackage{graphicx}
\def\sym#1{\ifmmode^{#1}\else\(^{#1}\)\fi}
\begin{document}

\title{Contrastive Learning of Relative Position Regression for One-Shot Object Localization in 3D Medical Images}

\author{Wenhui Lei\inst{1}\and Wei Xu\inst{1} \and Ran Gu\inst{1} \and Hao Fu\inst{1} \and Shaoting~Zhang\inst{1,2} \and Guotai~Wang\inst{1}\thanks{Correspond author.}}
\institute{
$^1$School of Mechanical and Electrical Engineering, University of Electronic Science and Technology of China, Chengdu, China\\$^2$SenseTime Research, Shanghai, China\\
\email{guotai.wang@uestc.edu.cn}\\
}
\maketitle
\footnote{First Author and Second Author contribute equally to this work.}
\begin{abstract}
Deep learning networks have shown promising performance for object localization in medial images, but require large amount of annotated data for supervised training. 
To address this problem, we propose: 1) A novel contrastive learning method which embeds the anatomical structure by predicting the Relative Position Regression (RPR) between any two patches from the same volume; 2) An one-shot framework for organ and landmark localization in volumetric medical images. 
Our main idea comes from that tissues and organs from different human bodies own similar relative position and context. Therefore, we could predict the relative positions of their non-local patches, thus locate the target organ.  
Our one-shot localization framework is composed of three parts: 1) A deep network trained to project the input patch into a 3D latent vector, representing its anatomical position;
2) A coarse-to-fine framework contains two projection networks, providing more accurate localization of the target; 3) Based on the coarse-to-fine model, we transfer the organ bounding-box (B-box) detection to locating six extreme points along x, y and z directions in the query volume. Experiments on multi-organ localization from head-and-neck (HaN) and abdominal CT volumes showed that our method acquired competitive performance in real time, which is more accurate and $10^5$ times faster than template matching methods with the same setting for one-shot localization in 3D medical images. Code is public at \href{https://github.com/LWHYC/RPR-Loc}{https://github.com/LWHYC/RPR-Loc}. 
\end{abstract}

\section{Introduction}
\par Localization of organs and anatomical landmarks is crucial in medical image analysis, as it can assist in the treatment, diagnosis and follow-up of many diseases\cite{zhang2017detecting}. Nowadays, Deep Learning has achieved state-of-the-art performance in the localization of a broad of structures, including abdominal organs\cite{xu2019efficient}, brain tissues\cite{vlontzos2019multiple}, etc. However, their success mainly relies on a large set of annotated images for training, which is expensive and time-consuming to acquire.
\par To reducing the demand on human annotations, several techniques such as weakly-supervised-\cite{oquab2015object}, semi-supervised-\cite{wang2020focalmix}, self-supervised-\cite{wang2020lt,zhou2019models,jing2020self} and one-shot learning (OSL) \cite{fei2006one, bart2005cross} attracted increasing attentions. OSL is especially appealing as it does not require annotations of the target during training, and only needs one annotated support image at the inference time. However, to the best of our knowledge, there have been very few works on one-shot object localization in volumetric medical images, such as Computed Tomography (CT) and Magnetic Resonance Images (MRI).
\par 
In this paper, we propose a one-shot localization method for organ/landmark localization in volumetric medical images, which does not require annotations of either the target organs or other types of objects during training and can directly locate any landmarks or organs that are specified by a support image in the inference stage. To achieve this goal, we present a novel contrastive learning\cite{oord2018representation, tian2019contrastive,hjelm2018learning} method called Relative Position Regression (RPR). Our main inspiration comes from the fact that the spatial distributions of organs (landmarks) have strong similarities among patients. Therefore, we could project every part of scans to a shared 3D latent coordinate system by training a projection network (Pnet) to predict the 3D physical offset between any two patches from the same image. We represent the bounding box of a target object by six extreme points, and locate these extreme points respectively by using Pnet to predict the offset from an agent's current position to the target position. Finally, we propose a coarse-to-fine framework based on Pnet that allows an agent to move mulitple steps for accurate localization, and introduce a multi-run ensemble strategy to further improve the performance. Experiments on multi-organ localization from head-and-neck (HaN) and abdominal CT volumes showed that our method achieved competitive performance in real time. Therefore, our method removes the need of human annotations for training deep learning models for object localization in 3D medical images. 

\par 
\section{Methodology}

\subsubsection{Relative Position Regression and Projection Network} 
Let $\bm{v}$ denote a volumetric image in the unannotated training set, and let $\bm{x_q}$ and $\bm{x_s}$ represent a moving (query) patch and a reference (support) patch with the same size $D\times H \times W$ in $\bm{v}$ respectively, we use a network to predict the 3D offset from $\bm{x_q}$ to $\bm{x_s}$. Assuming the pixel spacing of $\bm{v}$ is $\bm{e}\in R^3$ while $\bm{c_{q}}, \bm{c_{s}}\in R^3$ represent the centroid coordinates of $\bm{x_q}$ and $\bm{x_s}$, the ground truth offset $\bm{d'_{qs}}$ from $\bm{x_q}$ to $\bm{x_s}$ in the physical space is denoted as:
\begin{equation}
    \bm{d'_{qs}}= (\bm{c_{s}}-\bm{c_{q}}) \circ \bm{e}
\end{equation}
where $\circ$ represents the element-wise product.
\par
\begin{figure}[ht]
\vspace{-0.6cm}
    \centering
    \includegraphics[width=0.6\linewidth]{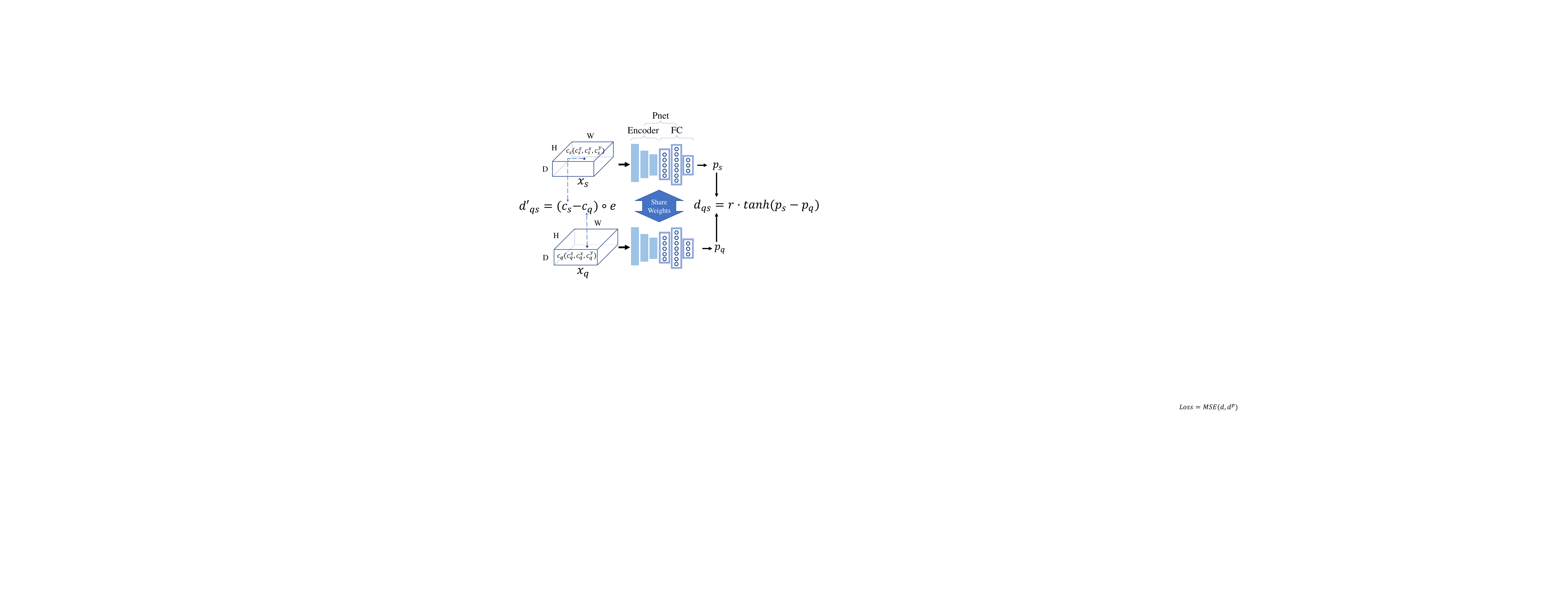}
    \caption{Structure of our projection network (Pnet) for offset prediction. $\bm{x_s}$ and $\bm{x_q}$ are the support and query patches centered at $\bm{c_s}$ and $\bm{c_q}$. We use a shared Pnet to transform $\bm{x_s}$ and $\bm{x_q}$ to 3D latent vectors $\bm{p_s}$ and $\bm{p_q}$, respectively. The Pnet contains convolution blocks to extract features and fully connected layers for projection. We apply scale factor $r$ and hyperbolic tangent function $tanh$ to get the predicted offset $\bm{d_{qs}}$, i.e., relative position from $\bm{x_s}$ to $\bm{x_q}$.}
    \label{fig:projection}
    \vspace{-0.2cm}
\end{figure}

 As shown in Fig.~\ref{fig:projection}, our Pnet is composed of two parts: convolution blocks to extract high-level features and fully connected layers mapping the features to a 3D latent vector. The details could be found in supplementary material. With Pnet, $\bm{x_s}$ and $\bm{x_q}$ are projected into $\bm{p_s}$ and $\bm{p_q}$ $\in R^3$. Then the predicted offset $\bm{d_{qs}} \in R^3$ from the query patch $\bm{x_q}$  to the support patch $\bm{x_s}$ is obtained as:
\begin{equation}
    \bm{d_{qs}}= r \cdot tanh(\bm{p_{s}}-\bm{p_{q}})
\end{equation}
where the hyperbolic tangent function $tanh$ and the hyper-parameter $r$ together control the upper and lower bound of $\bm{d_{qs}}$, which is set to cover the largest possible offset. Finally, we apply the mean square error (MSE) loss function to measure the difference between $\bm{d_{qs}}$ and $\bm{d'_{qs}}$:
\begin{equation}
    Loss = ||\bm{d_{qs}}- \bm{d'_{qs}}||^2
\end{equation}

\par In the inference stage for landmark localization, we use an annotated support (reference) image to specify the desired landmark, and aim to find the position of the corresponding landmark in a test image. 
The problem is modeled as moving an agent from a random initial position to the target position. Therefore, we take a patch at a random position $\bm{c_0}\in R^3$ in the test image as $\bm{x_q}$, which is the initial status of the agent, and take the patch centered at the specified landmark in the support image as the support patch, which is an approximation of the unknown target patch in the test image and serves as $\bm{x_s}$. By applying Pnet with $\bm{x_s}$ and $\bm{x_q}$, we obtain an offset $\bm{d_{qs}}\in R^3$, thus we move the agent with $\bm{d_{qs}}$, and obtain $\bm{c=c_0 + d_{qs}}$ as the detected landmark position in the test image. 
\subsubsection{Multi-step localization}
As moving the agent with only one step to the target position may not be accurate enough, our framework supports multi-step localization, i.e., Pnet can be applied several times given the agent's new position and the support patch. In this paper, we employ two-step inference strategy as that showed the best performance during experiment. The first step obtains a coarse localization, where the offset can be very large. The second step obtains a fine localization with a small offset around the coarse localization result.
\par
\begin{figure}[h!]
    \vspace{-0.5cm}
    \centering
    \includegraphics[width=1\linewidth]{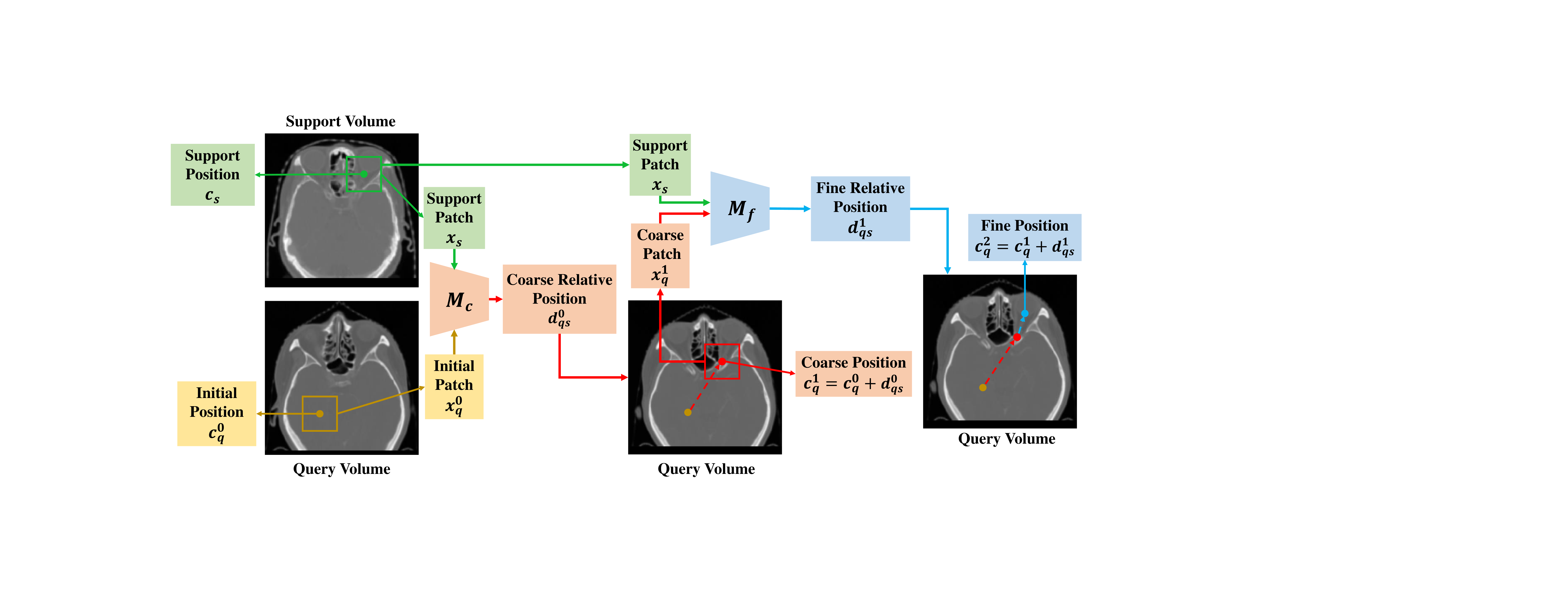}
    \caption{Proposed one-shot localization method based on Relative Position Regression (RPR-Loc). The green box represents a support (reference) patch in the support image that specifies a landmark to locate in a test image, and the yellow box is an initial patch that represents the current position of an agent in the test image. $\bm{M_c}$ (coarse projection model) takes the support patch and initial patch as input to regress the offset from the initial patch to the desired landmark in the test image. The agent moves one step given by the offset, and its new position is represented by the red box, which is a coarse result of the localization. We apply the $\bm{M_f}$ (fine projection model) to obtain a fine localization result given the agent's new position and the support patch. }
    \label{fig:framework}
    \vspace{-0.2cm}
\end{figure}

 To further improve the performance, we train two models for these two steps respectively. More specifically, for the coarse model $\bm{M_c}$, we crop $\bm{x_q}$ and $\bm{x_s}$ randomly across the entire image space and set $r_c$ to cover the largest possible offset between $\bm{x_q}$ and $\bm{x_s}$. For the fine model $\bm{M_f}$, we first randomly crop the $\bm{x_q}$, then crop $\bm{x_s}$ around $\bm{x_q}$ within a small range of $r_f$ mm, e.g., 30 mm. Therefore, $\bm{M_c}$ could handle large movements while $\bm{M_f}$ then focuses on the movement with small steps. During the inference stage, we use $\bm{M_c}$ to predict the coarse position, then use $\bm{M_f}$ to obtain the fine position. The RPR-Loc framework is shown in Fig.\ref{fig:framework}.


\vspace{-0.8cm}
\subsubsection{Multi-run ensemble}
As the position of the agent during inference is initialized randomly, different initialization may obtain slightly different results. To obtain more robust results, we propose a multi-run ensemble strategy: we locate each target landmark $K$ times with different random initializations, and average the localization results.

\subsubsection{Organ Detection via Landmark Localization}
\begin{figure}[ht]
    \vspace{-0.4cm}
    \centering
    \includegraphics[width=0.9\linewidth]{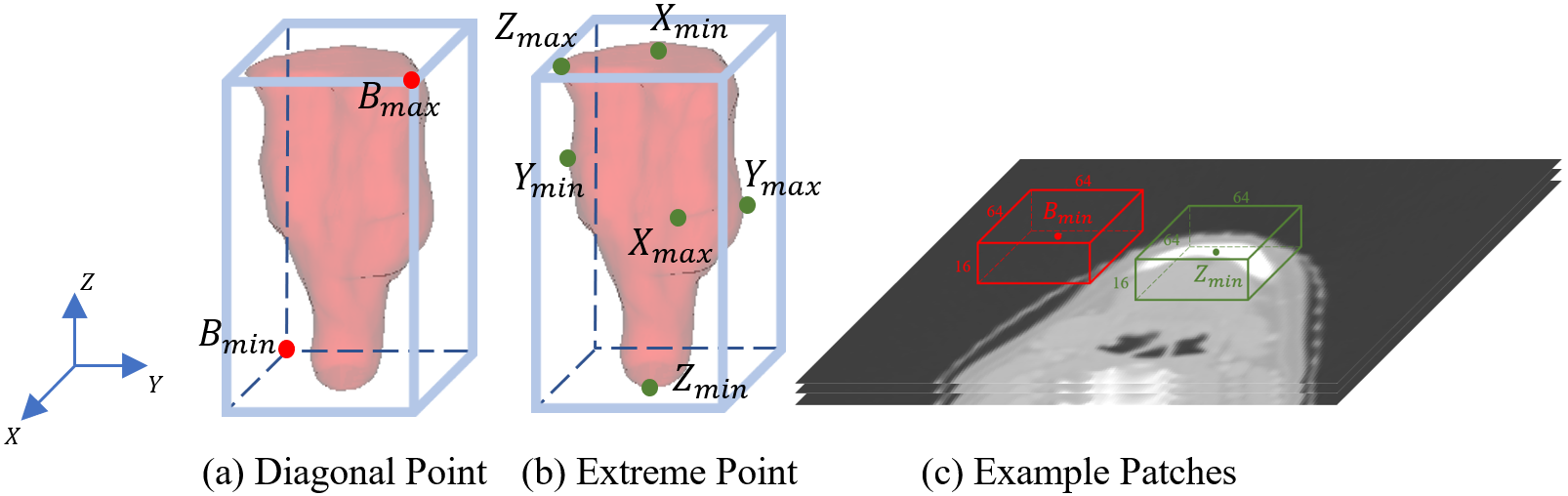}
    \caption{Example of two methods transferring B-box estimation to landmark localization: (a) Directly locating the bottom-left corner and top-right corner of B-box; (b) Finding 6 extreme points along x, y, and z axes; (c) An example of the two different types of points and their surrounding patches of the mandible. The patch around $D_{min}$ (minimum diagonal point) locates at meaningless area (air), while the one around $Z_{min}$ (minimum extreme point along axial orientation) links directly to the bound of the mandible.}
    \label{fig:extreme_point}
    \vspace{-0.2cm}
\end{figure}

We have described how to use Pnet to locate a landmark, and it can be easily reused for locating the bounding box of an organ without additional training. Specifically, for the organ B-box prediction, we transfer it to a set of landmark localization problems. As shown in Fig. \ref{fig:extreme_point}, there are two possible methods to implement this. The first is to locate the maximum and minimum coordinates of B-box among three dimensions by predicting bottom-left corner and top-right corner $[\bm{D}_{min}, \bm{D}_{max}]$ \cite{law2018cornernet}, which is referred as "Diagonal Points" in Fig. \ref{fig:extreme_point}(a). However, these two points may be located in meaningless areas (e.g., air background). As the case shown in Fig. \ref{fig:extreme_point}(c), the minimum diagonal point $D_{min}$ of the mandible is hard to detect due to lack of contextual information. Therefore, linking patches directly with the wanted organs would be more reasonable, and we propose to locate the six "Extreme Points" on the surface of an organ along x, y and z axes, which are denoted as $[\bm{Z}_{min}, \bm{Z}_{max}, \bm{X}_{min}, \bm{X}_{max}, \bm{Y}_{min}, \bm{Y}_{max}]$ in Fig. \ref{fig:extreme_point}(b). As shown in Fig. \ref{fig:extreme_point}(c), using  the extreme points can ensure these landmarks be located in the body foreground, thus more contextual information can be leveraged to obtain more accurate position and scale of the associated B-box. 

\vspace{-0.6cm}
\section{Experiments}

\subsubsection{Dataset and Evaluation Metrics}
Our method was evaluated on two datasets: 1) A mixed head and neck (HaN) CT dataset containing 165 volumes from three sources: 50 patients from StructSeg 2019\footnote{https://structseg2019.grand-challenge.org/Home/}, 48 patients from MICCAI 2015 Head and Neck challenge~\cite{raudaschl2017evaluation} and 67 patients collected locally. We used four organs that were annotated in all of them: brain stem, mandible, left and right parotid gland. 
The spacing was resampled to $3\times1\times1$ mm along axial, coronal, sagittal directions and cropped to remove the air background. The 165 volumes were splitted into 109 training, 19 validation and 38 testing samples; 2) A pancreas-CT dataset\cite{roth2015deeporgan} with 82 volumes. The spacing was resampled to $3\times3\times3$ mm. We split the data into 50, 11 and 21 for training, validation and testing, respectively.
\par The ground truth 3D B-boxes and extreme points in the validation and testing images were inferred from the 3D masks of each organ. While during the inference stage, we randomly select one image from the validation set as the support (reference) image, and use each image in the testing set as the query image, respectively. For the comparison of different methods, we calculated two metrics: (1) the Intersection over Union (IoU) between the predicted and ground-truth B-box; (2) the Absolute Wall Distance (AWD) following~\cite{xu2019efficient,article}. 

\subsubsection{Implementation Details}
Our model was trained  with an NVIDIA GTX 1080 Ti GPU. The Adam optimizer~\cite{kingma2014adam} was used for training with batch size 6, initial learning rate $10^{-3}$ and 250 epochs. With regard to every query volume, we constrain each initial position in the non-air region. Fixed patch sizes of 16$\times$64$\times$64 and 48$\times$48$\times$48 pixels were used for the HaN and pancreas organ detection problem. The coarse and fine model $\bm{M_c}$, $\bm{M_f}$ are trained separately. We set $r_c=300, r_f=30$ for HaN and $r_c=700, r_f=50$ for pancreas.


\subsubsection{Performance of organ localization and multi-run ensemble}
We first compare the two different landmark representation strategies: "Diagonal Points" and "Extreme Points" and also investigate the effect of hyper-parameter $K$ for multi-run ensemble on the localization performance. As shown in Table~\ref{tab:point_patch}, with one initial patch, predicting extreme points rather than diagonal points can prominently increase IoU value and reduce AWD for all organs, especially for the mandible. Because in different volumes, the diagonal points on the bounding box may be in featureless regions such as the air background and difficult to locate, thus the performance of B-box prediction is limited in this strategy. Therefore, we apply the "extreme points" location strategy for all subsequent experiments. 
 
 \par Last five rows of Table~\ref{tab:point_patch} show that the average accuracy will also increase obviously as the number of runs for multi-run ensemble increases from 1 to 20. This is in line with our expectations that taking average of predictions from different start points would increase the precision and robustness.
 
\begin{table}[]
\vspace{-0.3cm}
\caption{The mean organ localization IoU (\%, left) and AWD (mm, right) of four HaN OARs and pancreas under different strategies and different numbers of multi-run ensemble (MRE) with our coarse model. "Diagonal" and "Extreme" mean organ detection via localization of diagonal points and extreme points.}
\label{tab:point_patch}
\scalebox{0.9}{
\begin{tabular}{@{}cc|ccccc|c@{}}
\toprule
strategy          & MRE     & Brain stem          & Mandible            & L Parotid           & R Parotid           & Mean  & Pancreas   \\ \midrule
Diagonal          & 1       & 44.5, 7.06          & 32.3, 15.29         & 36.5, 9.94          & 35.7, 9.66          & 37.2, 10.49 & 43.3, 14.51 \\
Extreme           & 1       & 45.9, 6.61          & 62.2, 8.04          & 38.6, 9.61          & 37.5, 9.09          & 46.1, 8.34 & 44.4, 13.85 \\
Extreme           & 5       & 49.4, 5.63          & 62.7, 7.52          & 41.8, 8.66          & 36.8, 9.13          & 47.7, 7.73 & 47.2, 12.78 \\
Extreme           & 10      & 50.8, 5.34          & 63.2, 7.40          & 42.4, 8.73          & 38.4, 8.88          & 48.7, 7.59 & 45.2, 13.53\\
Extreme           & 15      & 50.7, 5.34          & \textbf{63.7, 7.28} & \textbf{43.1, 8.48} & \textbf{38.8, 8.79} & \textbf{49.1, 7.48} & 46.0, 13.19 \\
Extreme           & 20      & \textbf{51.9, 5.16} & 63.3, 7.45          & 42.6, 8.57          & 37.4, 9.03          & 48.8, 7.55 & \textbf{48.1, 12.42}\\ \midrule
\end{tabular}}
\vspace{-1cm}
\end{table}

\subsubsection{Performance of multi-step localization}
Based on the "Extreme Points" strategy for organ detection, we report the effectness of our multi-step model in Table~\ref{tab:coarse-fine} with $K = 15$ for HaN and $K=20$ for pancreas. Moving the agent by two or three steps with coarse model $\bm{M_c}$ than just one step does not guarantee performance improvement. Because $\bm{M_c}$ is trained with a large $r_c$ thus owning a sparse latent space, leading to its insensitivity about small step offset. In contrast, using $\bm{M_c}$ in the first step followed by using $\bm{M_f}$ in the second step achieved the best performance among HaN and pancreas tasks. It shows that the fine model $\bm{M_f}$ trained within small offset range $r_f$ could deal with the movements of small steps more effectively.
\begin{table}[]
\vspace{-0.3cm}
\caption{The mean localization IoU (\%, left) and AWD (mm, right) of four HaN OARs and pancreas of different multi-step methods.}
\label{tab:coarse-fine}
\scalebox{0.9}{
\begin{tabular}{@{}c|ccccc|c@{}}
\toprule
Method                 & Brain stem          & Mandible            & L Parotid           & R Parotid           & Mean   &Pancreas             \\ \midrule
$\bm{M_c}$ (one step)       & 50.7, 5.34          & 63.7, 7.28          & 43.1, 8.48          & 38.8, 8.79          & 49.1, 7.48 & 48.1, 12.42  \\
$\bm{M_c}$ (two steps)      & 51.4, 5.30          & 68.0, 6.03          & 45.0, 8.23          & 38.7, 8.89          & 50.7, 7.11 & 46.7, 13.46 \\
$\bm{M_c}$ (three steps)    & 50.5, 5.52          & 68.7, 5.79          & \textbf{45.3}, 8.30 & 38.9, 8.81          & 50.9, 7.10 & 46.0, 13.76 \\
$\bm{M_c}$ \textbf{+} $\bm{M_f}$ & \textbf{61.5, 3.70} & \textbf{70.0, 5.39} & 44.8, \textbf{7.74} & \textbf{44.2, 7.65} & \textbf{55.1, 6.12} & \textbf{49.5, 12.22} \\ \midrule
\end{tabular}}
\vspace{-1cm}
\end{table}

\subsubsection{Comparison with other methods}
In this section, we compare our method with: 1) A state-of-the-art supervised localization method Retina U-Net\cite{jaeger2020retina} that was trained with the ground truth bounding boxes in the training set. 2) Template matching-based alternatives for one-shot organ localization under the same setting as our method. We crop patches around each extreme point from the support volume, then implement template matching method by sliding window operation with stride 2 along each dimension to find the patch in the test image that is the most similar to the support patch. We consider the following similarity-based methods for comparison: a) Gray Scale MSE (GS MSE), where mean square error of gray scale intensities is used for similarity measurement between the support and query patches; b) Gray Scale (GS Cosine), which means using cosine similarity as criteria for comparison; c) Gray Scale Normalized Cross Correlation (GS NCC). We use normalized cross correlation to evaluate the similarity; d) Feature Map MSE (FM MSE). We train an auto-encoder network \cite{lei2019deepigeos} to transform a patch into a latent feature, and use MSE to measure the similarity between the latent features of $\bm{x_q}$ and $\bm{x_s}$; e) Feature Map Cosine (FM Cosine), where cosine similarity is applied to the latent features of $\bm{x_q}$ and $\bm{x_s}$.



With the same query volume, Table~\ref{tab:other-method} shows comparison of these methods on accuracy and average time consumption for each organ. Our method outperforms GS MSE, GS Cosine, GS NCC and FM MSE by an average IoU score and AWD. Despite FM Cosine slightly outperformed our method in terms of IoU of the right parotids, its performance is much lower on the brain stem, and need to list the scores. In addition, our method is $1.7\times10^5$ times faster than FM cosine that uses sliding window scanning. Compared with the fully supervised Retina U-Net \cite{jaeger2020retina}, our one-shot method still have a gap, but the difference is small for the mandible, and our method is much faster than Retina U-Net \cite{jaeger2020retina}. 

\begin{table}[]
\vspace{-0.2cm}
\caption{Quantitative evaluation of different methods for 3D organ localization. For each organ, we measured the mean IoU (\%, left) and AWD (mm, right).  Note that Retina U-Net \cite{jaeger2020retina} requires full annotations, while the others does not need annotations for training.}
\label{tab:other-method}
\scalebox{0.8}{
\begin{tabular}{@{}c|ccccc|c|c@{}}
\toprule
Method    & Brain stem          & Mandible            & L Parotid           & R Parotid           & Mean                & Pancreas        & Time(s)       \\ \midrule
Ours      & \textbf{61.5, 3.70} & \textbf{70.0, 5.39} & \textbf{44.8, 7.74} & 44.2, \textbf{7.65} & \textbf{55.1, 6.12} & \textbf{49.5, 12.22}     & \textbf{0.15}\\ 
GS MSE \cite{bankman1993optimal}    & 39.3, 7.47          & 65.7, 7.20          & 39.8, 10.67         & 37.1, 10.67         & 45.5, 9.00          & 1.8, 90.23      & 1052\\ 
GS Cosine \cite{liu2019pedestrian} & 35.2, 7.83          & 67.4, 7.10          & 36.9, 11.07         & 36.7, 10.30         & 44.1, 9.08          & 2.7, 118.14     & 1421\\
GS NCC \cite{briechle2001template}   & 46.9, 7.35          & 58.9, 9.76          & 24.3, 18.78         & 23.9, 20.14         & 38.5, 14.01         & 3.5, 101.77     & 4547\\
FM MSE\cite{zou2012deep}    & 44.7, 6.52          & 67.7, 6.88          & 39.2, 10.43         & 40.6, 8.79          & 48.1, 8.16          & 7.9, 120.36     & 26586 \\
FM Cosine\cite{bodla2017deep} & 50.2, 5.91          & 69.1, 5.53          & 44.7, 7.97          & \textbf{44.6}, 8.22 & 52.2, 7.92          & 19.2, 56.27     & 25976 \\ \midrule
Retina U-Net\cite{jaeger2020retina} & 68.9, 3.54      & 75.1, 6.17          & 60.5, 5.83          & 63.9, 4.75 & 67.1, 5.07      & 81.0, 3.23  & 4.7 \\
\bottomrule
\end{tabular}}
\end{table}

\begin{figure}[ht]
 \vspace{-0.2cm}
    \centering
    \includegraphics[width=1\linewidth]{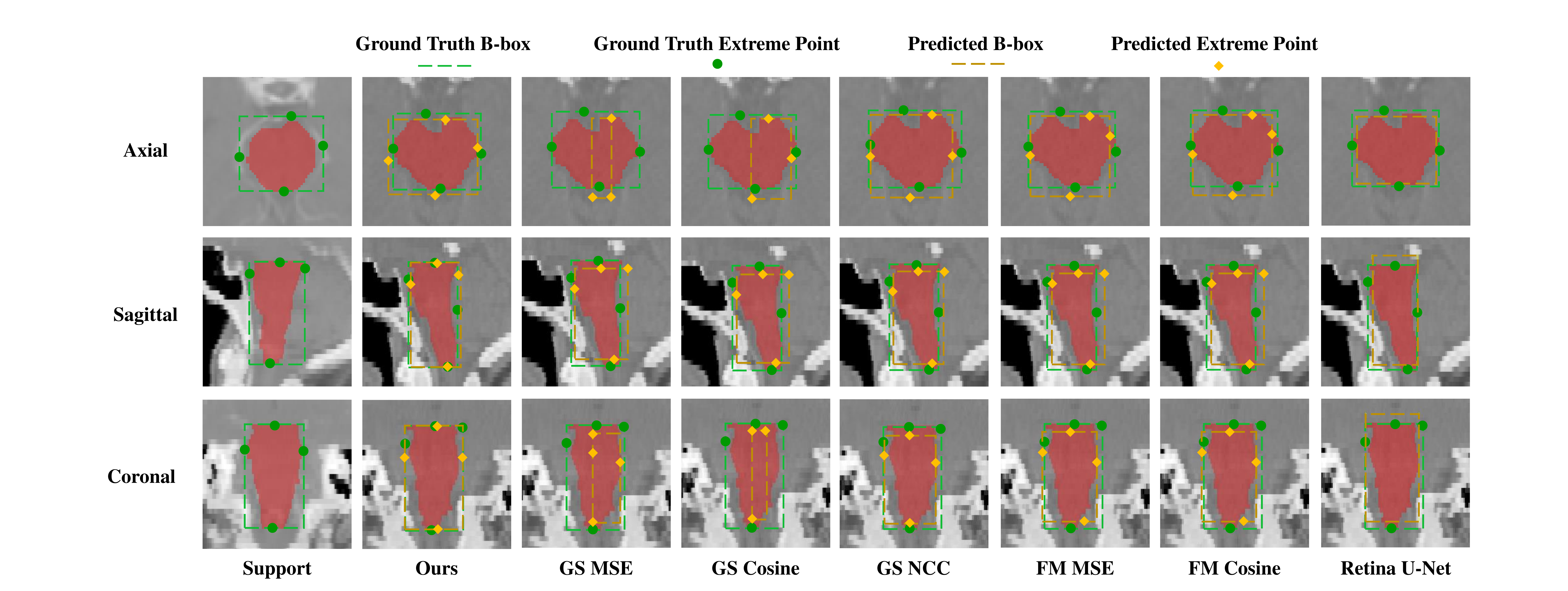}
    \caption{Qualitative localization results of brain stem in a query volume on three dimensions. The proposed method achieves desirable results.}
    \label{fig:vis}
\end{figure}

As shown in Fig. \ref{fig:vis}, despite the large variation in shapes and low contrast to surrounding tissues, the proposed framework could accurately locate B-box in the query volume. This implies that our method is promising for removing the need of annotation for training deep learning models for object localization in 3D medical images.

\section{Discussion \& Conclusions}
In this work, we propose a relative position regression-based one-shot localization framework (RPR-Loc) for 3D medical images, which to our best knowledge is the first work of one-shot localization in volumetric medical scans. Note that our one-shot localization can be easily extended to few-shot localization given multiple support volumes. Opposed to the traditional time-consuming template matching methods, our framework is regression-based thus not sensitive to the volume size and more efficient. Our method does not need any annotation during the training stage and could be employed to locate any landmarks or organs contained in the training dataset during the inference stage. Results on multi-organ localization from HaN and pancreas CT volumes showed that our method achieved more accurate results and is thousand to million times faster than template matching methods under the same setting.  This study demonstrates the effectiveness of our RPR-Loc in avoiding annotations of training images for deep learning-based object detection in 3D medical images, and it is of interest to apply it to other organs in the future. 


\bibliographystyle{splncs04}
\bibliography{MICCAI19.bib}
\end{document}